\documentclass{article}

\usepackage{microtype}
\usepackage{graphicx}
\usepackage{subcaption}
\usepackage{booktabs} 
\usepackage{multirow}
\usepackage{hyperref}

\usepackage[accepted]{icml2026}

\usepackage{amsmath}
\usepackage{amssymb}
\usepackage{mathtools}
\usepackage{amsthm}
\usepackage{caption}
\usepackage[table]{xcolor}  
\usepackage[capitalize,noabbrev]{cleveref}

\theoremstyle{plain}

\theoremstyle{definition}

\theoremstyle{remark}

\usepackage[textsize=tiny]{todonotes}

\icmltitlerunning{TIMI: Training-Free Image-to-3D Multi-Instance Generation with Spatial Fidelity}

\begin{document}

\twocolumn[

   \icmltitle{TIMI: Training-Free Image-to-3D Multi-Instance Generation \\ with Spatial Fidelity}



  \icmlsetsymbol{equal}{*}

  \begin{icmlauthorlist}
    \icmlauthor{Xiao Cai}{xxx}
    \icmlauthor{Pengpeng Zeng}{yyy}
    \icmlauthor{Ji Zhang}{sch}
    \icmlauthor{Heng Tao Shen}{yyy}
    \icmlauthor{Jingkuan Song}{yyy,comp}
    \icmlauthor{Lianli Gao}{xxx}
  \end{icmlauthorlist}

  \icmlaffiliation{xxx}{School of Computer Science and Engineering, University of Electronic Science and Technology of China, Chengdu, China}
  \icmlaffiliation{yyy}{School of Computer Science and Technology, Tongji University, Shanghai, China}
  \icmlaffiliation{comp}{Shanghai Innovation Institute, Shanghai, China}
  \icmlaffiliation{sch}{School of Computing and Artificial Intelligence, Southwest Jiaotong University, Chengdu, China}

  \icmlcorrespondingauthor{Pengpeng Zeng}{is.pengpengzeng@gmail.com}

  \icmlkeywords{Computer Vision, ICML}

  \vskip 0.3in

  \begin{center}
    {\includegraphics[width=0.9\textwidth]{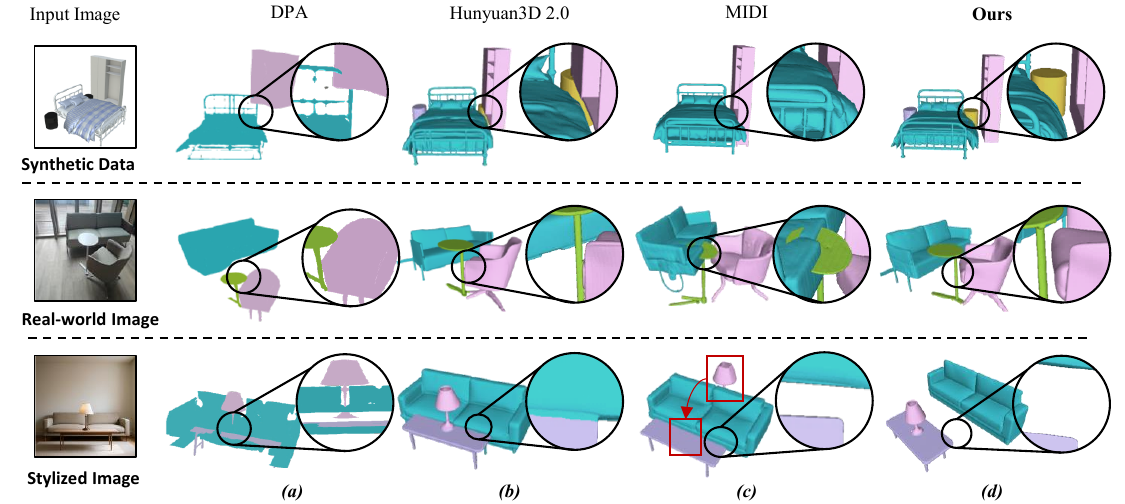}}
    \captionof{figure}{\textbf{Comparison of our training-free multi-instance 3D generation method with existing approaches.} TIMI generates multiple 3D instances from a single image by training-free guiding pre-trained 3D diffusion models, achieving precise global layout and distinct local instances, without requiring additional fine-tuning.}
    \label{fig:intro}
  \end{center}
]



\printAffiliationsAndNotice{}  


\begin{abstract}

Precise spatial fidelity in Image-to-3D multi-instance generation is critical for downstream real-world applications. Recent work attempts to address this by fine-tuning pre-trained Image-to-3D (I23D) models on multi-instance datasets, which incurs substantial training overhead and  struggles to guarantee spatial fidelity. 
In fact, we observe that pre-trained I23D models already possess meaningful spatial priors, which remain underutilized as evidenced by instance entanglement issues. 
Motivated by this, we propose \textbf{\textit{TIMI}}, a novel \textbf{T}raining-free framework for \textbf{I}mage-to-3D \textbf{M}ulti-\textbf{I}nstance generation that achieves high spatial fidelity. 
Specifically, we first introduce an Instance-aware Separation Guidance (ISG) module, which facilitates instance disentanglement during the early denoising stage. Next, to stabilize the guidance introduced by ISG, we devise a Spatial-stabilized Geometry-adaptive Update (SGU) module that promotes the preservation of the geometric characteristics of instances while maintaining their relative relationships. Extensive experiments demonstrate that our method yields better performance in terms of both global layout and distinct local instances compared to existing multi-instance methods, without requiring additional training and with faster inference speed.

\end{abstract}

\section{Introduction}

Image-to-3D multi-instance (I2MI) generation~\cite{midi,nie2020total3dunderstanding,DPA} seeks to synthesize 3D scenes containing multiple instances from a single image, serving as a foundation for engineering, product design, and creative industries. In contrast to the rapid progress in single-instance generation~\cite{liu2024one, voleti2024sv3d, shuang2025direct3d, liang2024luciddreamer, ProlificDreamer}, I2MI remains challenging and underexplored due to its stringent spatial fidelity requirements, which demand both accurate global layout and distinct local instances.

Recent breakthroughs in diffusion models~\cite{dit} have significantly propelled Image-to-3D (I23D) generation, establishing strong 3D priors that enable high-quality geometry synthesis from a single image. Building upon these advances, early I2MI methods adopt a compositional paradigm~\cite{chen2024comboverse, DPA, sing3d}, where individual instances are first generated independently using pre-trained I23D models and then assembled into a shared 3D scene through additional optimization or layout refinement. While straightforward, such multi-stage pipelines are prone to error accumulation, often resulting in inconsistent global layouts, spatial drift, or inter-instance collisions, as illustrated in Fig.~\ref{fig:intro}(a). To overcome these limitations, recent methods such as MIDI~\cite{midi} propose fine-tuning pre-trained I23D models with specialized attention, modeling inter-instance spatial relationships directly within the 3D generation process.

Despite their effectiveness, training-based approaches inevitably introduce substantial training overhead and still fail to fully guarantee spatial fidelity, such as imprecise layouts and fused instances, as shown in Fig.~\ref{fig:intro}(c).
Notably, we observe that pre-trained I23D models already possess meaningful spatial priors. As evidenced by Fig.~\ref{fig:intro}(b), Hunyuan3D 2.0~\cite{hunyuan3d} can often maintain reasonable inter-instance spatial relationships, but still exhibits instance entanglement. These observations naturally raise the following question: 
\textbf{\textit{Is it possible to exploit and disentangle the spatial priors of pre-trained I23D models for high spatial fidelity I2MI generation in a more flexible manner?}}

Inspired by this, we propose \textbf{\textit{TIMI}}, a novel \textbf{T}raining-free framework for \textbf{I}mage-to-3D \textbf{M}ulti-\textbf{I}nstance generation.
Specifically, TIMI introduces an {Instance-aware Separation Guidance} (ISG) module, which performs \emph{instance-aware attention anchoring} aligning early cross-attention with image-derived instance regions, and promotes instance disentanglement through an instance-consistent separation loss.
To stabilize this guidance, we further design a {Spatial-stabilized Geometry-adaptive Update} (SGU) module, which applies \emph{spatial-stabilized regularization} to smooth separation gradients and \emph{geometry-adaptive modulation} to control update magnitudes, thereby helping preserve geometric structure and layout during denoising.
Extensive experiments across diverse datasets demonstrate that our method consistently achieves superior global layout accuracy and more distinct local instances (Fig.~\ref{fig:intro}(d)) compared to existing I2MI methods, while requiring no additional training and enabling faster inference (Tab.~\ref{tab:main_results}).
Our contributions are threefold:
\begin{itemize}
\item We propose \textbf{TIMI}, a novel \textbf{training-free} framework for image-to-3D multi-instance generation with high spatial fidelity.
\item We introduce ISG and SGU to jointly promote local instance separation and global layout preservation.
\item Extensive experiments demonstrate the superiority of TIMI in both global and local spatial fidelity, achieving state-of-the-art layout alignment and instance distinctiveness without any additional training.
\end{itemize}

\section{Related Works}

\subsection{Single-instance 3D Generation}
The evolution of single-instance 3D generation reveals a clear transition from optimization-based pipelines to native 3D generation. Early approaches largely relied on Score Distillation Sampling (SDS)~\cite{dreamfusion, magic3d, chen2023fantasia3d} to distill 2D priors into 3D representations, but often suffered from slow optimization and geometric ambiguities. To enhance efficiency and consistency, subsequent research explored multi-view diffusion~\cite{zero123, mvdream, dreamview} and feed-forward reconstruction frameworks~\cite{lgm, xu2024instantmesh, zou2024triplane}. 
Most recently, diffusion-based 3D generation has further advanced feed-forward 3D creation by leveraging diffusion models~\cite{ldm, dit} to directly generate 3D-aware latent representations, enabling high-quality geometry and texture synthesis~\cite{3dtopia-xl, semv3d, trellis, hunyuan3d}.
These methods achieve impressive fidelity and efficient inference for single-object generation.
However, they are primarily designed for isolated object generation and still struggle to generalize to multi-instance scenarios without additional training.


\subsection{Multi-instance 3D Generation}
Multi-instance 3D generation has progressed from traditional reconstruction~\cite{chu2023buol, liu2022towards, paschalidou2021atiss} and retrieval-based methods~\cite{gao2024diffcad, gumeli2022roca, kuo2021patch2cad} to current generative frameworks. Recent efforts mainly fall into two categories: compositional generation and training-based approaches.
Compositional methods~\cite{tang2024diffuscene,DPA,rahamim2024lay, sam3d} typically follow a ``decompose-and-assemble'' strategy, where individual instances are generated independently and subsequently composed into a scene using auxiliary cues such as depth, layout, or canonical alignment~\cite{gen3dsr,sing3d,reparo,cast}. While flexible, these multi-stage pipelines are often computationally expensive and prone to error accumulation due to the lack of global scene reasoning.
In contrast, recent training-based methods~\cite{midi} pursue an end-to-end solution by fine-tuning I23D models with multi-instance attention to capture inter-instance spatial interactions. Despite their effectiveness, this paradigm incurs substantial training overhead and may weaken learned spatial priors.  In light of this, we exploit the strong priors of pre-trained I23D models through training-free instance-aware guidance to enable multi-instance generation with high spatial fidelity.

\section{Preliminary: 3D Object Generation Models}
\label{sec:preliminary}

Our work builds upon large-scale Image-to-3D generation frameworks~\cite{hunyuan3d, trellis}. These models typically comprise three core components: 
(1) A {3D Variational Autoencoder (3D VAE)} consisting of an encoder $\mathcal{E}$ and a decoder $\mathcal{D}$, which compress 3D geometric representations into a low-dimensional latent space, $z_0 \in \mathbb{R}^{L \times d}$, where $L$ denotes the sequence length and $d$ is the feature dimension.
(2) {Condition Encoders}, which extract semantic feature sequences $c \in \mathbb{R}^{M \times d}$ from the reference image $I$ using pre-trained vision models (e.g., CLIP~\cite{clip} or DINOv2~\cite{dino}).
(3) A {Denoising Transformer} $\epsilon_\theta$, trained to progressively recover $z_0$ from Gaussian noise $z_T \sim \mathcal{N}(0, \mathbf{I})$.

\noindent\textbf{Unified Joint Attention Mechanism.} 
Modern DiT architectures facilitate multi-modal interaction through a {joint attention} mechanism. Let $z_t$ denote the 3D latent tokens and $c$ denote the image condition tokens. By concatenating them into a joint sequence $U = [c; z_t] \in \mathbb{R}^{(M+L) \times d}$, the interaction is modeled by a global self-attention matrix $\mathbf{A}_{\text{global}}$, which can be decomposed into block-wise components:
\begin{equation}
    \mathbf{A}_{\text{global}} = \text{Softmax}\left(\frac{\mathbf{Q}_{U}\mathbf{K}_{U}^\top}{\sqrt{d}}\right) = 
    \begin{bmatrix} 
    \mathbf{A}_{cc} & \mathbf{A}_{cz} \\ 
    \mathbf{A}_{zc} & \mathbf{A}_{zz} 
    \end{bmatrix}.
    \label{eq:joint_attn}
\end{equation}
Here, the sub-matrix $\mathbf{A}_{zc} \in \mathbb{R}^{L \times M}$ (the bottom-left block) represents the \textbf{3D-to-Image Cross-Attention}, governing how 3D geometric tokens query semantic information from the 2D reference, which is therefore the focus of our method.

\begin{figure*}
    \centering
    \includegraphics[width=0.9\linewidth]{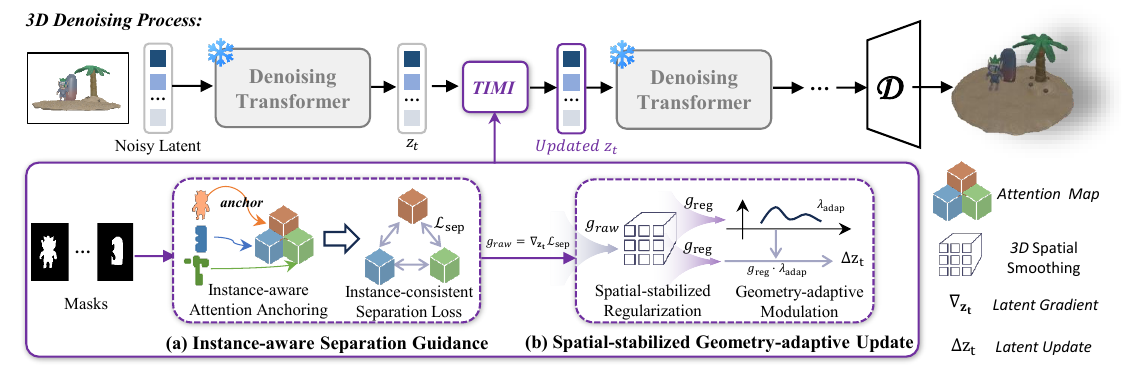}
    \caption{\textbf{Overview of the TIMI framework.} Given a single image and instance masks, TIMI guides a frozen pre-trained Image-to-3D diffusion model to generate multi-instance 3D outputs without  additional training. 
    \textbf{(a) Instance-aware Separation Guidance} applies instance-level constraints to early cross-attention layers to promote instance separation. 
    \textbf{(b) Spatial-stabilized Geometry-adaptive Update} stabilizes inference-time guidance via geometry-adaptive gradient modulation to preserve overall spatial structure.
    }
    \label{fig:pipeline}
\end{figure*}

\section{Method}

In this section, we present a training-free framework \textbf{\textit{TIMI}} for Image-to-3D multi-instance generation, which comprises two components: {Instance-aware Separation Guidance} (ISG) in Sec.~\ref{sec:separation} and {Spatial-stabilized Geometry-adaptive Update} (SGU) in Sec.~\ref{sec:update}, as shown in Fig.~\ref{fig:pipeline}.

\subsection{Instance-aware Separation Guidance}
\label{sec:separation}

To alleviate {instance entanglement}, we introduce the {Instance-aware Separation Guidance}, which promotes instance-level separation during the early denoising stages. 

\noindent\textbf{Instance-aware Attention Anchoring.}
To make instance-level spatial assignment explicit during Image-to-3D denoising, we operate on cross-attention maps in early layers, where 2D–3D correspondences are initially established.
In particular, we focus on the 3D-to-image cross-attention matrix $\mathbf{A}_{zc}$, which encodes dense spatial alignments between 3D latent tokens and 2D image features.
However, this raw attention remains instance-agnostic and may become ambiguous in multi-instance scenes, especially when different instances overlap spatially in the image.

To address this ambiguity, we introduce an \emph{instance-aware attention anchoring} operation that projects the attention distribution onto discrete semantic instances.
Given a set of instance masks $\mathcal{M} = \{M_k\}_{k=1}^K$ extracted from the reference image, we map $\mathbf{A}_{zc}$ to an instance-level representation by computing an \emph{Instance Probability Map} $\mathbf{P} \in \mathbb{R}^{L \times K}$:
\begin{equation}
\mathbf{P}{\cdot, k} = \mathbf{A}_{zc} \cdot M_k^\top.
\end{equation}
Here, $\mathbf{P}_{v,k}$ indicates the degree of association between the $v$-th 3D latent token and the $k$-th instance.
This projection provides an instance-aligned representation of 3D tokens, serving as the basis for subsequent instance-level separation.

\noindent\textbf{Instance-consistent Separation Loss.}
Based on the instance-aware representation $\mathbf{P}$, our objective is to guide the denoising process toward consistent instance separation in the 3D latent space.
Naively encouraging each 3D token to associate with a single instance may lead to degenerate solutions, such as attention collapse around instance centroids, compromising local structure.

To address this issue, we introduce an \emph{instance-consistent separation loss} that incorporates \emph{structure-aware spatial weighting}.
Specifically, we derive a spatial weight matrix $\mathbf{W}$ from the masked attention distribution, reflecting the relative spatial support of each token within a given instance:
\begin{equation}
\mathbf{W}_{v,k} = \frac{\mathbf{A}_{zc, v} \cdot M_k}{\sum_{v'} (\mathbf{A}_{zc, v'} \cdot M_k) + \epsilon}.
\end{equation}
The weight $\mathbf{W}_{v,k}$ emphasizes tokens that are structurally relevant to an instance, allowing the optimization to focus on meaningful spatial regions.
Using this instance-aware weighting, we formulate the separation objective as a spatially weighted negative log-likelihood:
\begin{equation}
\mathcal{L}_{\text{sep}} = - \sum_{k=1}^{K} \sum_{v=1}^{L} \mathbf{W}_{v,k} \log(\mathbf{P}_{v,k} + \epsilon).
\end{equation}
This loss promotes separation across instances while encouraging internal consistency within each instance, enabling early and stable instance disentanglement during denoising.

\subsection{Spatial-stabilized Geometry-adaptive Update}
\label{sec:update}
To stabilize instance-aware guidance while preserving coherent spatial structure, we propose the {Spatial-stabilized Geometry-adaptive Update}.

\paragraph{Spatial-stabilized Regularization (SR).}
The raw separation gradients $\nabla_{\mathbf{z}} \mathcal{L}_{sep}$ are often sparse and dominated by high-frequency components in the high-dimensional latent space.
Directly applying such gradients can disrupt spatial coherence, leading to unstable layouts or fragmented geometric structures.
To address this issue, we introduce a spatial-stabilized regularization (SR) on latent features.
Specifically, we apply an isotropic 3D Gaussian smoothing to the separation gradients prior to the update:
\begin{equation}
\label{eq_smoothing}
    \mathbf{g}_{reg} = \mathcal{K}_{\sigma} * \nabla_{\mathbf{z}} \mathcal{L}_{sep},
\end{equation}
where $\mathbf{z} \in \mathbb{R}^{C \times D \times H \times W}$ denotes the 3D latent features, $*$ is the 3D convolution operator, and $\mathcal{K}_{\sigma}$ is a Gaussian kernel with standard deviation $\sigma$.
This operation suppresses high-frequency perturbations and enforces spatially coherent gradient propagation, allowing separation guidance to act on contiguous latent regions and helping maintain local geometric continuity during inference-time optimization.

\begin{figure*}[t]
    \centering
    \includegraphics[width=0.85\linewidth]{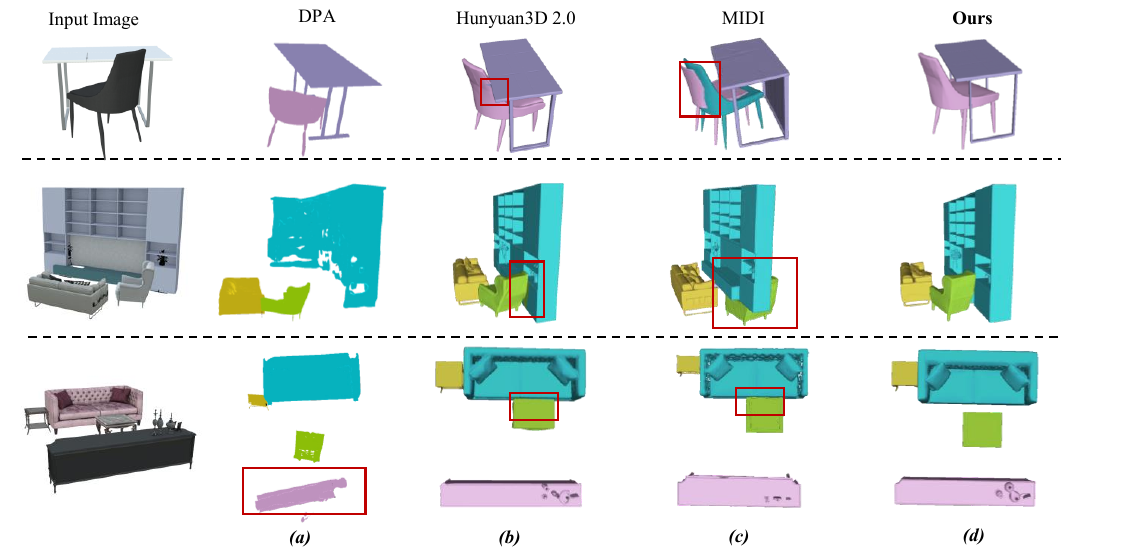}
    \caption{\textbf{Qualitative Comparisons on Synthetic Data.} Existing methods often produce inaccurate global layouts or fail to separate local instances. Our method preserves both global and local spatial fidelity, yielding well-disentangled instances. }
    \label{fig:main_com}
\end{figure*}

\paragraph{Geometry-adaptive Modulation (GM).}
Different instances exhibit significant variance in geometric sensitivity within the latent space. Applying a uniform update scale may cause excessive deformation for thin or fragile structures, while being insufficient for large or low-response instances.
To address this imbalance, we propose a geometry-adaptive gradient modulation strategy based on peak normalization.
At each denoising step $t$, we first compute the maximum magnitude of the regularized gradients: $\mu_{max}^{(t)} = \max |\mathbf{g}_{reg}^{(t)}|.$
Subsequently, we compute an adaptive scaling factor $\lambda_{adap}$ based on the statistical properties of the current latent feature distribution, ensuring that the maximum feature update is strictly controlled relative to the feature energy scale:
\begin{equation}
    \lambda_{adap} = \frac{\alpha \cdot \sigma_{\mathbf{z}_t}}{\mu_{max}^{(t)} + \epsilon},
    \label{eq:scale}
\end{equation}
where $\sigma_{\mathbf{z}_t}$ is the standard deviation of the current latent features $\mathbf{z}_t$, and $\epsilon$ is a stability term ($\epsilon$ = 1e-6) to prevent zero division. Here, $\alpha$ controls the maximum update magnitude relative to the current latent feature scale.

The final update vector $\Delta \mathbf{z}_t$ is then computed as $\Delta \mathbf{z}_t = \lambda_{adap} \cdot \mathbf{g}_{reg}^{(t)}$. By constraining the peak update magnitude, this mechanism balances optimization across heterogeneous geometric structures, preventing collapse in high-gradient regions while maintaining sufficient separation force elsewhere.
To further stabilize the optimization trajectory across timesteps, we apply a momentum-based update:
\begin{equation}
    \begin{aligned}
        \mathbf{m}_t &= \beta \mathbf{m}_{t-1} + (1-\beta)\Delta \mathbf{z}_t, \\
        \mathbf{z}_{t} &\leftarrow \mathbf{z}_{t} - \mathbf{m}_t,
    \end{aligned}
\end{equation}
where $\beta$ is the {momentum coefficient } (set to $0.9$),  which smooths temporal updates and mitigates oscillations during inference-time optimization.

Overall, the proposed Spatial-stabilized Geometry-adaptive Update transforms raw separation gradients into controlled, geometry-aware latent updates, enabling stable and effective local instance separation while maintaining global layout coherence throughout the denoising process.

\section{Experiments}
\subsection{Experimental Setup}

\paragraph{Implementation details}
We adopt Hunyuan3D 2.0~\cite{hunyuan3d} as the base pre-trained I23D model and follow its original inference pipeline and default configurations. Instance masks are extracted using Grounded-SAM~\cite{ren2024grounded}.  
ISG is applied to early cross-attention layers ($l \leq 4$) during early denoising steps ($t \leq 15$).  
For SGU, we set the gradient modulation strength to $\alpha = 0.1$ and the Gaussian smoothing standard deviation to $\sigma = 1.5$. 
We observe that these parameters work well across most cases, demonstrating the generalizability of TIMI. We also point out that better results may be obtained with a customized setting, e.g., a larger $\alpha$. 

\paragraph{Datasets.}
We evaluate multi-instance 3D generation on three datasets spanning diverse domains.
\textbf{(i) Synthetic Data:} We randomly sample $30$ test scenes from 3D-Front~\cite{3dfront}, each containing two or more spatially adjacent instances.
\textbf{(ii) Real-world Data:} To assess real-world generalization, we collect $20$ images from Real-Data~\cite{real-data}.
\textbf{(iii) Stylized Data:} To further evaluate cross-domain robustness, we generate $20$ stylized multi-instance images using FlUX.1 Kontext~\cite{flux}.

\paragraph{Baselines.}
We compare TIMI with representative I2MI approaches. Specifically, we include MIDI~\cite{midi}, a training-based method that performs supervised fine-tuning to enable instance disentanglement, and DPA~\cite{DPA}, a compositional approach that generates instances independently followed by scene-level composition. In addition, we directly compare against single-instance method Hunyuan3D 2.0~\cite{hunyuan3d}.

\paragraph{Metrics.}
We evaluate multi-instance 3D generation using complementary metrics that assess both global and local spatial fidelity.
Following prior works~\cite{nie2020total3dunderstanding, midi}, we report Chamfer Distance (CD) and F-Score (FS) at both global level (CD-S, FS-S) and local level (CD-O, FS-O).
To further measure layout alignment and instance disentanglement, we additionally adopt Layout Consistency Distance (LCD) and Separation Success Rate (SSR).
Detailed definitions and implementations of the metrics are provided in \textbf{Appendix~A}.

\begin{figure}[t]
    \centering
    \includegraphics[width=0.95\linewidth]{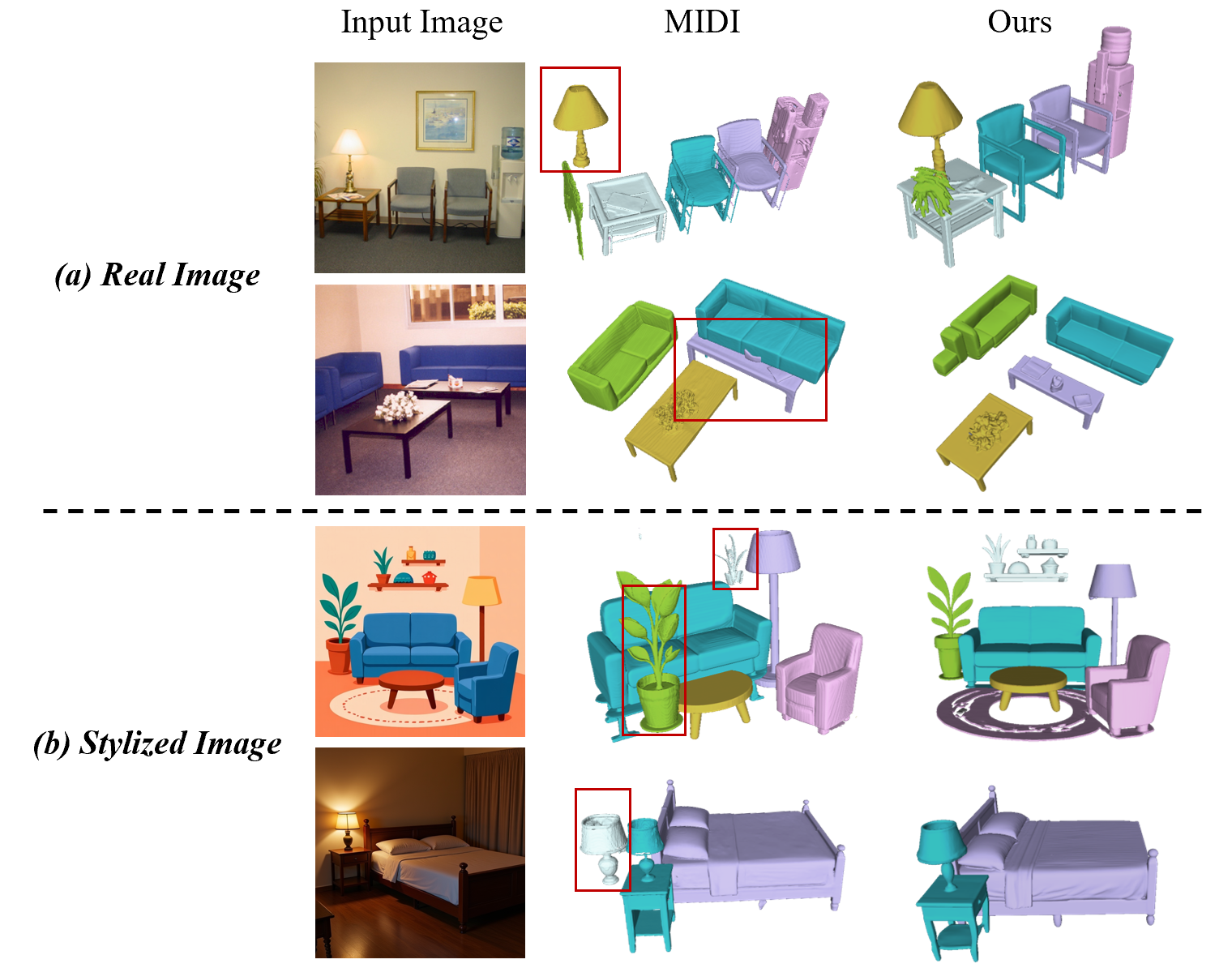}
    \caption{{\textbf{Qualitative comparisons on real and stylized images.}}
    TIMI preserves stronger global and local spatial fidelity across real-world and stylized inputs.
    }
    \label{fig:main_com_2}
\end{figure}

\begin{figure*}[t]
    \centering
    \includegraphics[width=0.85\linewidth]{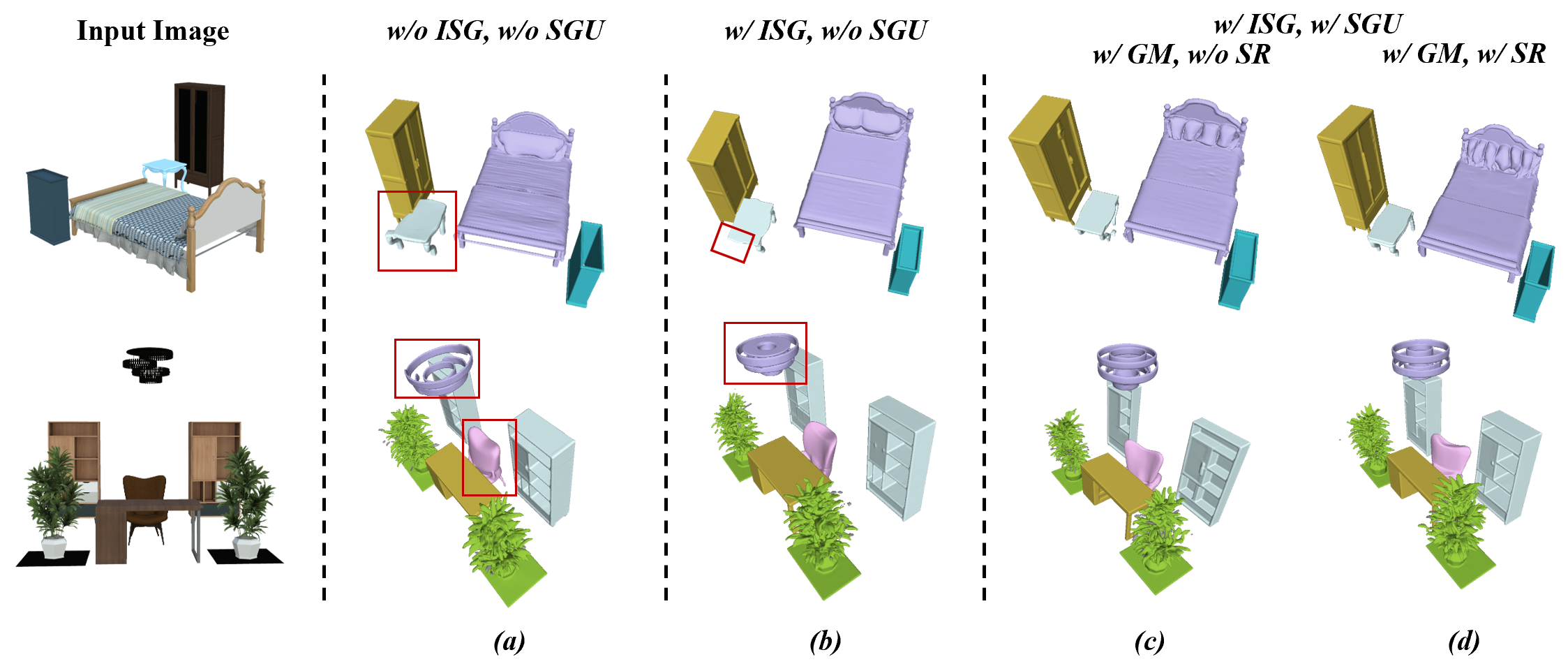}
    \caption{\textbf{Qualitative ablation study on the effectiveness of proposed components in TIMI.} The results demonstrate that each module progressively enhances the layout alignment and instance distinctiveness, with the full method achieving the best spatial fidelity.}
    \label{fig:ablation_component}
\end{figure*}

\subsection{Main Results}
\subsubsection{Qualitative Comparison}
To qualitatively evaluate the performance of TIMI, we present visual comparisons across diverse datasets.

\paragraph{I2MI Generation on Synthetic Data.} As illustrated in Fig.~\ref{fig:main_com}, TIMI consistently maintains global and local spatial fidelity across synthetic scenes. \textbf{(i) Global spatial fidelity.} TIMI faithfully preserves the spatial layout of the input. In contrast, MIDI exhibits noticeable layout drift, exemplified by the misaligned sofa and bookshelf in Fig.~\ref{fig:main_com}(c). \textbf{(ii) Local spatial fidelity.} At the instance level, TIMI generates structurally distinct instances with crisp boundaries, effectively overcoming the geometric fusion observed in Hunyuan3D 2.0 (e.g., the coalesced sofa and table in Fig.~\ref{fig:main_com}(b)). Collectively, these results confirm TIMI's capacity to maintain precise global layouts and distinct local instances.

\paragraph{I2MI Generation from Real and Stylized Images.}
We further extend our evaluation to real-world and stylized domains (Fig.~\ref{fig:main_com_2}), where TIMI continues to uphold high spatial fidelity.
Specifically, in terms of global layout, MIDI exhibits noticeable spatial drift, where instances such as plants are incorrectly placed in front of the sofa and table in Fig.~\ref{fig:main_com_2}(b), while our results remain well aligned with the input images. 
Regarding local instance, MIDI suffers from instance fusion, as evidenced by the overlapping table and sofa in Fig.~\ref{fig:main_com_2}(a), whereas TIMI preserves clear instance separation. These observations validate the effective generalization of our framework under varied inputs.

\begin{table}
    \centering
    \Large
    \caption{\textbf{Quantitative comparison on synthetic data.} Our method demonstrates superior global and local spatial fidelity while maintaining high inference efficiency.}
    \label{tab:main_results}
    \resizebox{1\linewidth}{!}{
        \begin{tabular}{c|c c c|c c c|c}
            \toprule
            \multirow{2}{*}{\textbf{Method}} & \multicolumn{3}{c|}{\textbf{Global Spatial Fidelity} } & \multicolumn{3}{c|}{\textbf{Local Spatial Fidelity} } & \textbf{Inference} \\
            
            \cmidrule(lr){2-4} \cmidrule(lr){5-7} \cmidrule(lr){8-8}
            
             & \textbf{LCD} $\downarrow$ & \textbf{CD-S} $\downarrow$ & \textbf{FS-S} $\uparrow$ & \textbf{SSR} $\uparrow$ &  \textbf{CD-O} $\downarrow$ & \textbf{FS-O} $\uparrow$ &\textbf{Time} $\downarrow$ \\
            \midrule
            
            Hunyuan3D 2.0  & 0.627 & 0.0492 & 0.450 & 0.697  & 0.0986 & 0.339 &\textbf{54.2s} \\
            DPA          & 0.649 & 0.0662 & 0.249 & 0.743 & 0.1620 & 0.124 & 783s \\
            MIDI & 0.634 & \textbf{0.0409} & 0.396 & 0.737 & \textbf{{0.0760}} & 0.312 & 90.1s\\
            \rowcolor{cyan!10}
            Ours   & \textbf{0.598} & \underline{0.0424} & \textbf{0.458} & \textbf{0.809}& \underline{0.0855} & \textbf{0.353} & \underline{59.2s} \\
            \bottomrule
        \end{tabular}
    }
\end{table}

\subsubsection{Quantitative Comparison}
To systematically evaluate TIMI, we conduct quantitative comparisons with existing methods, as summarized in Tab.~\ref{tab:main_results}. The results reveal that: 
\textbf{(i)} Despite being training-free, TIMI consistently outperforms the training-based approach MIDI in both global and local spatial fidelity. 
\textbf{(ii)} TIMI achieves the strongest overall global spatial fidelity, attaining the best LCD (\textbf{0.598}) and FS-S (\textbf{0.458}). Although MIDI yields a slightly lower CD-S, LCD more directly reflects layout alignment, indicating that TIMI reconstructs global layouts that are better aligned with the input image. 
\textbf{(iii)} For local spatial fidelity, TIMI substantially surpasses all baselines, achieving the highest SSR (\textbf{0.809}) and FS-O (\textbf{0.353}), demonstrating superior instance-level separation. While MIDI exhibits competitive CD-O, its lower SSR suggests more frequent instance fusion, whereas TIMI generates more distinct and well-separated instances. 
\textbf{(iv)} In addition, TIMI maintains high inference efficiency ($\sim$59s), comparable to the base model Hunyuan3D, and is significantly faster than both the training-based MIDI ($\sim$90s) and the compositional method DPA ($\sim$783s).
The quantitative evidence verifies that TIMI enhances both layout alignment and instance separation without sacrificing the efficiency of generation.

\begin{figure}[t]
    \centering
    \includegraphics[width=0.9\linewidth]{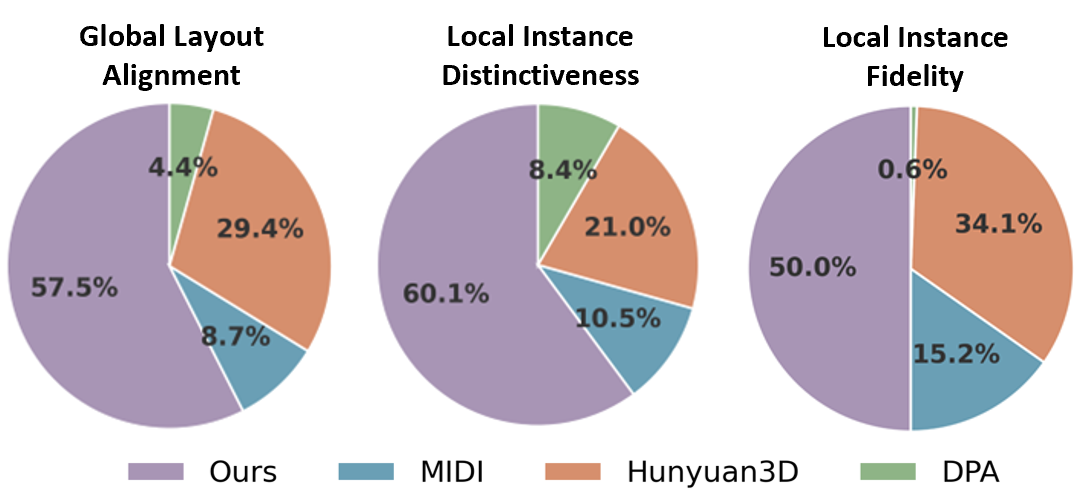}
    \caption{\textbf{{User study.}} Our method demonstrates higher subjective preference compared to other methods.}
    \label{fig:user_study}
\end{figure}

\paragraph{User Study.}
We further conduct a user study to assess perceptual quality under human evaluation. As summarized in Fig.~\ref{fig:user_study}, \textbf{(i)} 57.5\% of participants prefer our results in terms of global layout alignment. \textbf{(ii)} 60.1\% of users favor our method for local instance distinctiveness, significantly outperforming MIDI and Hunyuan3D, suggesting that fine-grained instance separation is more perceptually evident to human observers. \textbf{(iii)} Our method is also preferred by 50.0\% of users in terms of instance quality, indicating that improved global and local spatial fidelity is achieved without compromising generative quality.

\subsection{Ablation Study}
\label{sec:ablation}

In this section, we conduct comprehensive ablation studies to validate the effectiveness of components within our TIMI.
First, we verify the necessity of the two core modules: ISG and SGU.
Next, we provide a fine-grained analysis within ISG in Sec.~\ref{sec:ablation_ISG} and SGU in Sec.~\ref{sec:ablation_SGU}.

\paragraph{Effect of \textit{ISG} and $\textit{SGU}$.}
In this experiment, we investigate the effectiveness of ISG and SGU by progressively adding them to the baseline. From the results in Tab.~\ref{tab:ablation_components}, we have the following observations. \textbf{(i)} Both ISG and SGU contribute to the improvement of global and local spatial fidelity. \textbf{(ii) \textit{w/ ISG}.} Introducing the ISG effectively improves local spatial fidelity, as reflected by the gains in FS-O and CD-O. \textbf{(iii) \textit{w/ ISG, \textit{w/ SGU}}.} Further incorporating SGU substantially enhances global spatial fidelity while preserving the improved local instance quality, resulting in consistent gains across global and local metrics. 

These findings are further supported by qualitative comparisons in Fig.~\ref{fig:ablation_component}. 
\textbf{(i)} Without either module (Fig.~\ref{fig:ablation_component}(a)), instances such as the stool and cabinet exhibit severe fusion with the stool appearing incomplete and deformed. 
\textbf{(ii)} With ISG enabled via naive gradient updates (Fig.~\ref{fig:ablation_component}(b)), the stool becomes disentangled but suffers from geometric fractures (e.g., missing legs), confirming that direct gradient application disrupts 3D structural coherence.
\textbf{(iii)} By incorporating both ISG and SGU (Fig.~\ref{fig:ablation_component}(d)), the overall layout becomes more coherent, and the stool is correctly generated with complete structure and plausible geometry.

\begin{table}[t]
    \centering
    \caption{\textbf{Quantitative ablation study on components in TIMI,} progressively incorporating ISG and SGU effectively enhances the global and local spatial fidelity. }
    \label{tab:ablation_components}
    \resizebox{1\linewidth}{!}{
        \begin{tabular}{l|c c c|c c c}
            \toprule
            \multirow{2}{*}{} & \multicolumn{3}{c|}{\textbf{Global Spatial Fidelity} } & \multicolumn{3}{c}{\textbf{Local Spatial Fidelity} }  \\
            
            \cmidrule(lr){2-4} \cmidrule(lr){5-7} 
            
            & \textbf{LCD} $\downarrow$ & \textbf{CD-S} $\downarrow$ & \textbf{FS-S} $\uparrow$ & \textbf{SSR} $\uparrow$ & \textbf{CD-O} $\downarrow$ & \textbf{FS-O} $\uparrow$  \\
            \midrule
            
            w/o ISG, w/o SGU  &  0.627 & 0.0492 & 0.450 & 0.697  & 0.0986 & 0.339\\
            w/ ISG, w/o SGU  &  0.623 & 0.0462 & {0.448} & 0.674 & 0.0864  & {0.351} \\

            \rowcolor{cyan!10}
            w/ ISG, w/ SGU  & \textbf{{0.598}} & \textbf{{0.0424}} & \textbf{0.458} & \textbf{0.809}& \textbf{{0.0855}} & \textbf{{0.353}} \\
            
            \bottomrule
        \end{tabular}
    }
\end{table}

\subsubsection{Analysis of ISG}
\label{sec:ablation_ISG}

To further analyze the design choices of ISG, we conduct ablation studies on two key hyperparameters: (i) the selection of guided cross-attention layers ${l}$, and (ii) the number of early denoising timesteps $t$ over which ISG is applied.

\paragraph{Effect of $l$ in ISG.}
Tab.~\ref{tab:ablation_layer} reports the performance of ISG when applying guidance to different ranges of cross-attention layers. We summarize three key observations.
\textbf{(i)} Guiding the first four layers ($l \in [0,4]$) achieves the best overall trade-off, delivering strong global spatial fidelity (lowest CD-S), and the most effective instance disentanglement (highest SSR).
\textbf{(ii)} Restricting guidance to very shallow layers (e.g., $l \in [0,1]$) slightly improves FS-related metrics but results in degraded global spatial fidelity and weaker instance separation, indicating insufficient spatial regulation.
\textbf{(iii)} Increasing the layer coverage beyond the fourth layer (e.g., up to layer 8) brings no further improvement and instead results in a consistent performance drop compared to the four-layer setting. This indicates that deeper layers may capture higher-level semantics (e.g., texture), which are less effective for instance-level spatial regulation.
\begin{figure}[t]
    \centering
    \includegraphics[width=0.95\linewidth]{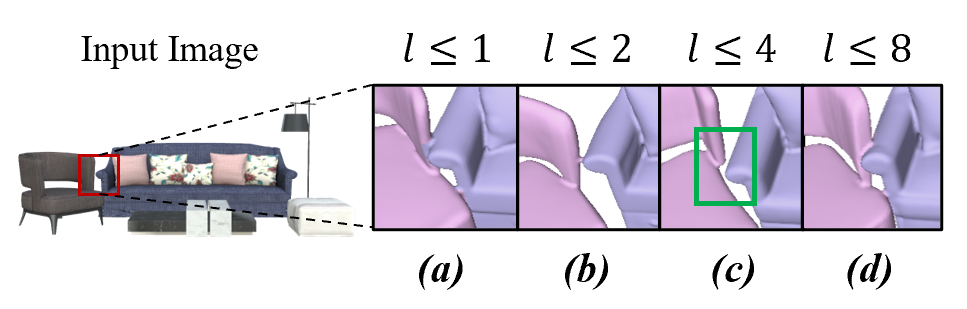}
    \caption{\textbf{Qualitative results of ablation study on $l$ in ISG.}
$l \le 4$ achieves the best instance separation, while smaller or larger $l$ leads to insufficient disentanglement or partial instance fusion.}
    \label{fig:abaltion_layers}
\end{figure}

\begin{table}[t]
    \centering
    \caption{\textbf{Quantitative results of ablation study on $l$ in ISG.}}
    \label{tab:ablation_layer}
    \resizebox{0.95\linewidth}{!}{
        \begin{tabular}{c|c c c|c c c}
            \toprule
            \multirow{2}{*}{\textbf{$l$}} & \multicolumn{3}{c|}{\textbf{Global Spatial Fidelity} } & \multicolumn{3}{c}{\textbf{Local Spatial Fidelity} }  \\
            
            \cmidrule(lr){2-4} \cmidrule(lr){5-7} 
            
            & \textbf{LCD} $\downarrow$ & \textbf{CD-S} $\downarrow$ & \textbf{FS-S} $\uparrow$ & \textbf{SSR} $\uparrow$ & \textbf{CD-O} $\downarrow$ & \textbf{FS-O} $\uparrow$  \\
            \midrule
            
            1  & 0.600 & 0.0437 & {0.454} & 0.772 & 0.0896  & \textbf{0.372} \\
            2  & 0.615 & 0.0443 & 0.453 & 0.785 & {0.0874} & 0.352 \\
            3  & \textbf{0.554} & 0.0449 & 0.439 & 0.797 & 0.1007 & 0.360 \\
            
            \rowcolor{cyan!10}
            4  & {0.598} & \textbf{0.0424} & \textbf{0.457} & \textbf{0.809} & \textbf{0.0855} & 0.353 \\
            
            5  & 0.594 & 0.0434 & 0.452 & 0.786 & 0.0938 & 0.359 \\
            
            8  & 0.616 & 0.0468 & 0.442 & 0.773 & 0.0931 & 0.358 \\
            \bottomrule
        \end{tabular}
    }
\end{table}

Fig.~\ref{fig:abaltion_layers} provides qualitative comparisons under different choices of guided cross-attention layers $l$.  
\textbf{(i)} When guidance is applied to the first four layers ($l \in [0,4]$), instance separation is the most visually clear and stable. As shown, the sofa armrest and the chair are cleanly disentangled with well-defined boundaries.
\textbf{(ii)} For shallower guidance ranges (e.g., $l \in [0,1]$ or $[0,2]$), noticeable instance entanglement is observed. In particular, structural parts of the sofa tend to merge with the nearby chair, indicating that guidance confined to very early layers is insufficient.
\textbf{(iii)} Extending guidance to deeper layers (e.g., $l \in [0,8]$) also degrades separation quality. Although the overall geometry remains plausible, object boundaries become less distinct, and partial fusion reappears between adjacent instances. This suggests that deeper layers, which may encode more abstract or semantic information, are less suitable for enforcing instance-level spatial separation.
Overall, the qualitative results confirm that guiding the first four layers achieves the best balance between instance disentanglement and structural preservation, in line with the quantitative performance.

\paragraph{Effect of $t$ in ISG.} 
We further analyze the effect of applying ISG at different denoising timesteps, with quantitative results summarized in Tab.~\ref{tab:ablation_steps}. 
\textbf{(i)} Applying ISG within the first 15 denoising steps achieves the most balanced performance, yielding the best global layout consistency (lowest CD-S) and the highest local instance distinctiveness (SSR), while maintaining competitive instance quality.
\textbf{(ii)} Limiting guidance to an overly short early duration ($t \le 10$) constrains instance disentanglement, as indicated by reduced SSR.
\textbf{(iii)} Conversely, extending ISG into later denoising stages ($t \ge 20$) brings no further gains in spatial fidelity and instead degrades instance separation (lower FS-O).

\begin{figure}[t]
    \centering
    \includegraphics[width=0.9\linewidth]{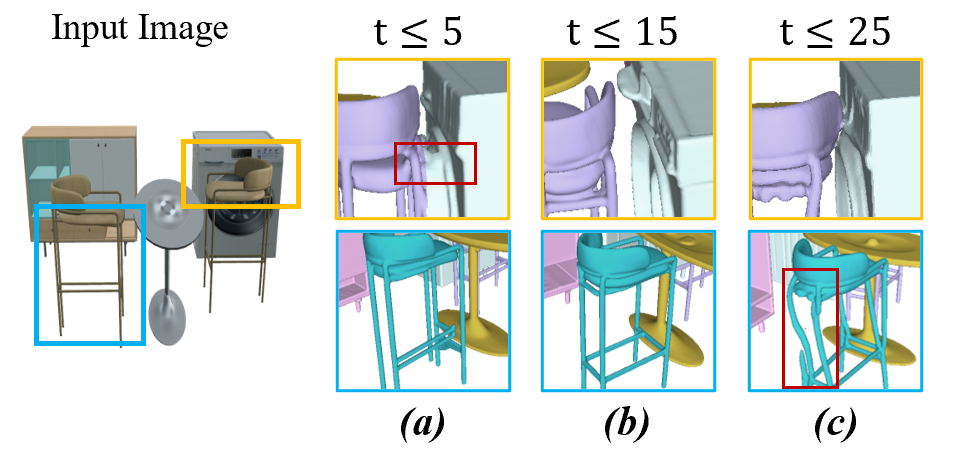}
    \caption{\textbf{Qualitative ablation study on $t$ in ISG.}
    $t \le 15$ yields the best trade-off, while shorter or longer schedules cause insufficient disentanglement or geometric distortion.
    }
    \label{fig:ablation_steps}
\end{figure}

These trends are visually supported in Fig.~\ref{fig:ablation_steps}. 
Applying ISG only at very early steps ($t \le 5$) results in insufficient separation, with the chair and washing machine remaining fused (Fig.~\ref{fig:ablation_steps}(a)), 
whereas overly prolonged guidance ($t \le 25$) introduces structural distortions, such as deformed chair legs (Fig.~\ref{fig:ablation_steps}(c)). 
Therefore, we select $t=15$ as the default setting, which provides a favorable balance between global layout alignment and local instance distinctiveness.

\begin{table}[t]
    \centering
    \caption{\textbf{Quantitative results of ablation study on $t$ in ISG.}}
    \label{tab:ablation_steps}
    \resizebox{1\linewidth}{!}{
        \begin{tabular}{c|c c c|c c c}
            \toprule
            \multirow{2}{*}{\textbf{$t$}} & \multicolumn{3}{c|}{\textbf{Global Spatial Fidelity} } & \multicolumn{3}{c}{\textbf{Local Spatial Fidelity} }  \\
            
            \cmidrule(lr){2-4} \cmidrule(lr){5-7} 
            
            & \textbf{LCD} $\downarrow$ & \textbf{CD-S} $\downarrow$ & \textbf{FS-S} $\uparrow$ & \textbf{SSR} $\uparrow$ & \textbf{CD-O} $\downarrow$ & \textbf{FS-O} $\uparrow$  \\
            \midrule
            
            5  & 0.554 & 0.0448 & {0.446} & 0.753 & 0.1041  & 0.353 \\
            10  & 0.563 & 0.0435 & 0.453 & 0.739 & {0.0859} & \textbf{0.383} \\

            \rowcolor{cyan!10}
            15  & {0.598} & \textbf{{0.0424}} & \textbf{0.458} & \textbf{0.809}& \textbf{{0.0855}} & \underline{0.353} \\
            
            20  & {0.603} & 0.0443 & 0.437 & 0.790 & 0.0991 & 0.335 \\
            
            25  & \textbf{0.552} & 0.0433 & {0.442} & 0.771 & 0.0976 & 0.342 \\
            \bottomrule
        \end{tabular}
    }
\end{table}

\subsubsection{Analysis of SGU}
\label{sec:ablation_SGU}
We conduct a detailed analysis of the SGU. Specifically, we first evaluate the individual contributions of its components, and then examine the effects of the guidance strength $\alpha$ in Eq.~\ref{eq:scale} and the spatial regularization factor $\sigma$ in Eq.~\ref{eq_smoothing}.

\paragraph{Effect of \textit{GM} and \textit{SR} in SGU.}
Tab.~\ref{tab:ablation_SGU} reports the ablation results of the individual components in SGU.
\textbf{(i)} Without GM and SR, the model exhibits inferior global spatial fidelity and weak instance disentanglement, as reflected by higher CD-S and the lowest SSR.
\textbf{(ii)} Introducing GM alone substantially improves global layout consistency (best LCD), but provides limited gains in instance disentanglement and local spatial fidelity.
\textbf{(iii)} Combining GM with SR yields consistent improvements across all metrics, achieving the best global and local spatial fidelity.

As shown in Fig.~\ref{fig:ablation_component}, with GM introduced into SGU, the model achieves a more stable and coherent global layout. However, noticeable instance-level structural distortions emerge, such as the twisted stool legs in Fig.~\ref{fig:ablation_component}(c). After further incorporating SR, the global layout remains stable while such distortions are effectively alleviated, enabling robust local instance separation.

\begin{table}[t]
    \centering
    \caption{\textbf{Quantitative ablation study on components in SGU,} showing that GM and SR jointly enhance global and local spatial fidelity.}
    \label{tab:ablation_SGU}
    \resizebox{1\linewidth}{!}{
        \begin{tabular}{l|c c c|c c c}
            \toprule
            \multirow{2}{*}{} & \multicolumn{3}{c|}{\textbf{Global Spatial Fidelity} } & \multicolumn{3}{c}{\textbf{Local Spatial Fidelity} }  \\
            
            \cmidrule(lr){2-4} \cmidrule(lr){5-7} 
            
            & \textbf{LCD} $\downarrow$ & \textbf{CD-S} $\downarrow$ & \textbf{FS-S} $\uparrow$ & \textbf{SSR} $\uparrow$ & \textbf{CD-O} $\downarrow$ & \textbf{FS-O} $\uparrow$  \\
            \midrule
            
            w/o GM, w/o SR  & 0.623 & 0.0462 & {0.448} & 0.674 & 0.0864  & {0.351}\\
            w/ GM, w/o SR  & \textbf{{0.597}} & 0.0455 & 0.444 & 0.764 & {0.0882} & \textbf{{0.368}} \\

            \rowcolor{cyan!10}
            w/ GM, w/ SR  & \underline{0.598} & \textbf{{0.0424}} & \textbf{0.458} & \textbf{0.809}& \textbf{{0.0855}} & \underline{0.353} \\
            
            \bottomrule
        \end{tabular}
    }
\end{table}

\begin{figure}[t]
    \centering
    \includegraphics[width=0.9\linewidth]{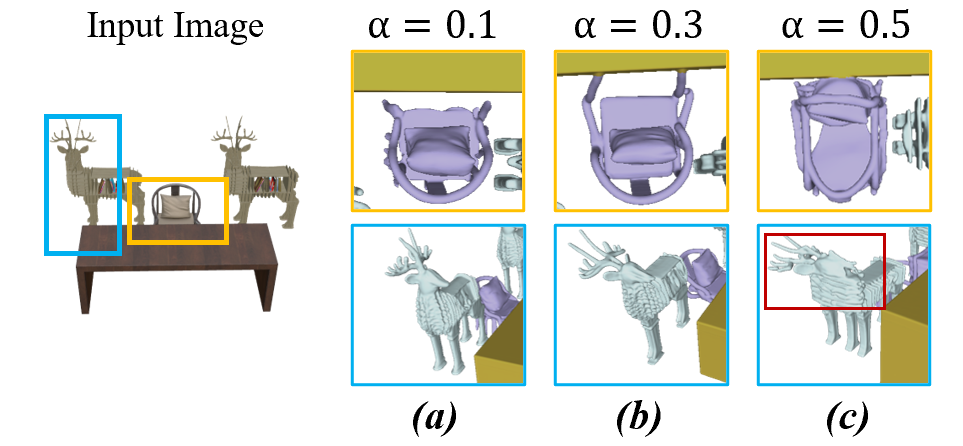}
    \caption{\textbf{Qualitative results of ablation study on $\alpha$ in SGU.} A moderate guidance strength ($\alpha=0.1$) yields the most balanced global layout consistency and instance-level geometric fidelity. }
    \label{fig:ablation_alpha}
\end{figure}

\begin{table}[t]
    \centering
    \caption{\textbf{Quantitative results of ablation study on $\alpha$ in SGU.} }
    \label{tab:ablation_alpha}
    \resizebox{1\linewidth}{!}{
        \begin{tabular}{c|c c c|c c c}
            \toprule
            \multirow{2}{*}{\textbf{${\alpha}$}} & \multicolumn{3}{c|}{\textbf{Global Spatial Fidelity} } & \multicolumn{3}{c}{\textbf{Local Spatial Fidelity} }  \\
            
            \cmidrule(lr){2-4} \cmidrule(lr){5-7} 
            
            & \textbf{LCD} $\downarrow$ & \textbf{CD-S} $\downarrow$ & \textbf{FS-S} $\uparrow$ & \textbf{SSR} $\uparrow$ & \textbf{CD-O} $\downarrow$ & \textbf{FS-O} $\uparrow$  \\
            \midrule
            \rowcolor{cyan!10}
            0.1  & {0.598} & \underline{0.0424} & \textbf{0.458} & \textbf{0.809}& \underline{0.0855} & \underline{0.353} \\
            0.2  & \textbf{0.523} & 0.0439 & 0.446 & 0.780 & {0.1001} & \textbf{{0.359}} \\

            
            0.3  & {0.562} & 0.0449 & 0.441 & {0.796} & 0.0897 & 0.326 \\
            
            0.4  & {0.611} & 0.0428 & 0.436 & 0.766 & 0.0862 & 0.315 \\
            
            0.5  & 0.570 & \textbf{0.0414} & {0.429} & 0.750 & \textbf{0.0755} & 0.314 \\
            \bottomrule
        \end{tabular}
    }
\end{table}

\paragraph{Effect of $\alpha$ in SGU.}
We investigate the sensitivity of the global guidance strength $\alpha$, with detailed results provided in Tab.~\ref{tab:ablation_alpha}. 
\textbf{(i)} When $\alpha$ is set to a small value (e.g., $\alpha=0.1$), SGU generally achieves strong global spatial fidelity while preserving robust local instance quality.
\textbf{(ii)} Slightly increasing $\alpha$ (e.g., $\alpha=0.2$) maintains comparable global spatial fidelity, but local instance quality begins to degrade, as reflected by increased instance-level CD.
\textbf{(iii)} As $\alpha$ is further increased to intermediate ranges (e.g., $0.3$ or $0.4$), both global and local spatial fidelity consistently deteriorate.
\textbf{(iv)} For large values of $\alpha$ (e.g., $\alpha=0.5$), although global alignment may remain strong, excessive guidance tends to destabilize instance geometry (degraded FS metrics).

These trends are further illustrated in Fig.~\ref{fig:ablation_alpha}. With smaller $\alpha$, instances exhibit more complete structures and clearer boundaries, whereas larger $\alpha$ values progressively introduce geometric degradation and instance interference, indicating that overly strong guidance can disrupt geometric stability and compromise spatial fidelity.

\paragraph{Effect of ${\sigma}$ in SGU.}
We analyze the influence of ${\sigma}$ in Eq.~\ref{eq_smoothing}, with quantitative results reported in Tab.~\ref{tab:ablation_sigma}.
We observe that a small ${\sigma}$ (e.g., $0.5$) leads to inferior global spatial fidelity and weaker instance-level separation, indicating insufficient spatial regularization. Conversely, an overly large ${\sigma}$ (e.g., $2.5$) degrades both global and local spatial fidelity, suggesting that overly strong smoothing may disrupt instance geometry. Consequently, we adopt ${\sigma}=1.5$ as the optimal setting, which achieves the best balance between layout consistency and instance distinctiveness (best CD-S and SSR). 

\begin{figure}[t]
    \centering
    \includegraphics[width=0.95\linewidth]{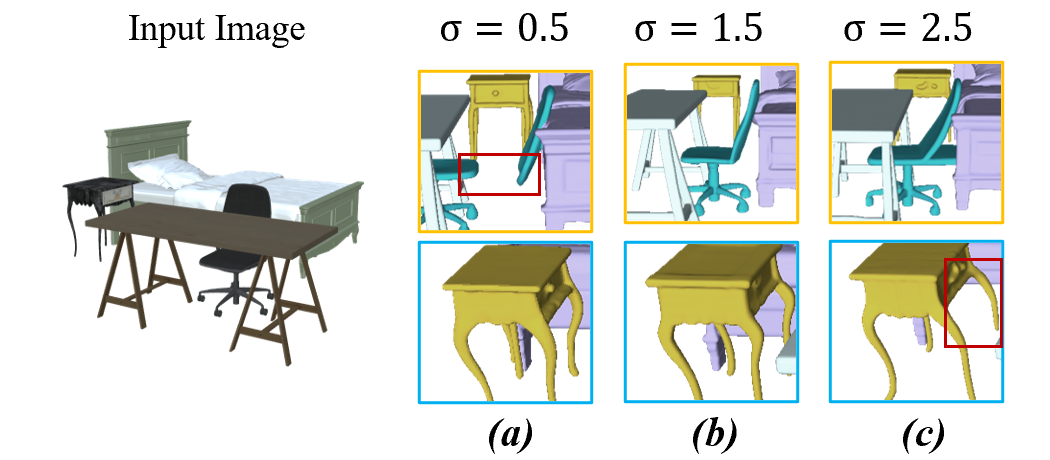}
    \caption{\textbf{Qualitative results of ablation study on $\sigma$ in SGU.}
$\sigma=1.5$ achieves the best balance between instance separation and geometric preservation, while smaller or larger $\sigma$ leads to insufficient disentanglement or structural distortion.} 
    \label{fig:ablation_sigma}
\end{figure}

We further conduct a qualitative comparison to investigate the impact of different $\sigma$ values, as visualized in Fig.~\ref{fig:ablation_sigma}. As shown in Fig.~\ref{fig:ablation_sigma}(a), an overly small $\sigma$ leads to unstable guidance, resulting in geometric fractures (e.g., the chair back is detached from its body) and insufficient disentanglement (e.g., the chair remains fused with the bed). This indicates that weak smoothing fails to enforce coherent separation forces. Conversely, as shown in Fig.~\ref{fig:ablation_sigma}(c), while an overly large $\sigma$ provides sufficient smoothing to separate the chair from the bed, it disrupts the intrinsic geometry of individual instances, causing noticeable structural distortions such as the twisted furniture leg. Based on these observations, we adopt $\sigma=1.5$ as the optimal default setting to balance separation effectiveness and geometric preservation.

\section{Conclusion}

We introduce TIMI, a training-free Image-to-3D multi-instance generation framework via instance-aware guidance. By exploiting the spatial priors of pre-trained I23D models, TIMI improves multi-instance generation without additional fine-tuning. The proposed ISG promotes local instance disentanglement during early denoising, while SGU stabilizes the guidance to preserve geometric structures and precise global layouts. Experiments across synthetic, real-world, and stylized inputs demonstrate that TIMI achieves superior global and local spatial fidelity over existing methods while maintaining efficient inference. These results suggest that TIMI offers a practical solution for high spatial fidelity multi-instance 3D generation. 

\begin{table}[t]
    \centering
    \caption{\textbf{Quantitative results of ablation study on ${\sigma}$ in SGU.} }
    \label{tab:ablation_sigma}
    \resizebox{1\linewidth}{!}{
        \begin{tabular}{c|c c c|c c c}
            \toprule
            \multirow{2}{*}{\textbf{${\sigma}$}} & \multicolumn{3}{c|}{\textbf{Global Spatial Fidelity} } & \multicolumn{3}{c}{\textbf{Local Spatial Fidelity} }  \\
            
            \cmidrule(lr){2-4} \cmidrule(lr){5-7} 
            
            & \textbf{LCD} $\downarrow$ & \textbf{CD-S} $\downarrow$ & \textbf{FS-S} $\uparrow$ & \textbf{SSR} $\uparrow$ & \textbf{CD-O} $\downarrow$ & \textbf{FS-O} $\uparrow$  \\
            \midrule
            
            0.5  & \textbf{0.556} & 0.0439 & {0.444} & 0.782 & 0.1057  & 0.352 \\
            1.0  & 0.622 & 0.0442 & 0.448 & 0.768 & \textbf{0.0837} & \textbf{0.379} \\

            \rowcolor{cyan!10}
            1.5  & {0.598} & \textbf{{0.0424}} & \textbf{0.458} & \textbf{0.809}& \underline{0.0855} & \underline{0.353} \\
            
            2.0  & {0.570} & 0.0428 & 0.444 & 0.784 & 0.0942 & 0.340 \\
            
            2.5  & {0.638} & 0.0436 & {0.440} & 0.782 & 0.1032 & 0.332 \\
            \bottomrule
        \end{tabular}
    }
\end{table}

\section*{Acknowledgments}
This study is supported by grants from Fundamental and Interdisciplinary Disciplines Breakthrough Plan of the Ministry of Education of China (No. JYB2025XDXM116), and the National Natural Science Foundation of China (Grant No. 62425208, No. U22A2097, No. U23A20315, No. 82441006).

\section*{Impact Statement}
This paper presents TIMI, a framework designed to advance 3D generative modeling by enabling high spatial fidelity multi-instance synthesis without the substantial computational costs of fine-tuning. By introducing a training-free paradigm, our work lowers the barrier to entry for high-quality 3D content creation, potentially democratizing access for applications in industrial design, virtual reality, and the creative industries. However, while our method significantly improves efficiency, it relies on pre-trained foundation models and may inherently reflect their underlying data biases. We do not foresee immediate negative societal consequences beyond general considerations regarding the responsible use of generative AI technologies.

\bibliography{example_paper}

@inproceedings{midi,
  title={Midi: Multi-instance diffusion for single image to 3d scene generation},
  author={Huang, Zehuan and Guo, Yuan-Chen and An, Xingqiao and Yang, Yunhan and Li, Yangguang and Zou, Zi-Xin and Liang, Ding and Liu, Xihui and Cao, Yan-Pei and Sheng, Lu},
  booktitle={IEEE/CVF Conference on Computer Vision and Pattern Recognition},
  pages={23646--23657},
  year={2025}
}

@inproceedings{real-data,
  title={Recognizing indoor scenes},
  author={Quattoni, Ariadna and Torralba, Antonio},
  booktitle={IEEE/CVF Conference on Computer Vision and Pattern Recognition},
  pages={413--420},
  year={2009}
}

@article{DPA,
  title={Zero-shot scene reconstruction from single images with deep prior assembly},
  author={Zhou, Junsheng and Liu, Yu-Shen and Han, Zhizhong},
  journal={Advances in Neural Information Processing Systems},
  volume={37},
  pages={39104--39127},
  year={2024}
}

@article{sam3d,
  title={Sam 3d: 3dfy anything in images},
  author={Chen, Xingyu and Chu, Fu-Jen and Gleize, Pierre and Liang, Kevin J and Sax, Alexander and Tang, Hao and Wang, Weiyao and Guo, Michelle and Hardin, Thibaut and Li, Xiang and others},
  journal={arXiv preprint arXiv:2511.16624},
  year={2025}
}

@article{semv3d,
  title={Semv-3d: Towards concurrency of semantic and multi-view consistency in general text-to-3d generation},
  author={Cai, Xiao and Zeng, Pengpeng and Gao, Lianli and Su, Sitong and Shen, Heng Tao and Song, Jingkuan},
  journal={IEEE Transactions on Image Processing},
  year={2026},
  publisher={IEEE}
}

@inproceedings{ldm,
  title={High-resolution image synthesis with latent diffusion models},
  author={Rombach, Robin and Blattmann, Andreas and Lorenz, Dominik and Esser, Patrick and Ommer, Bj{\"o}rn},
  booktitle={Proceedings of the IEEE/CVF conference on computer vision and pattern recognition},
  pages={10684--10695},
  year={2022}
}

@inproceedings{dit,
  title={Scalable diffusion models with transformers},
  author={Peebles, William and Xie, Saining},
  booktitle={IEEE/CVF Conference on Computer Vision and Pattern Recognition},
  pages={4195--4205},
  year={2023}
}

@inproceedings{dreamfusion,
  author       = {Ben Poole and
                  Ajay Jain and
                  Jonathan T. Barron and
                  Ben Mildenhall},
  title        = {DreamFusion: Text-to-3D using 2D Diffusion},
  booktitle    = {International Conference on Learning Representations},
  year         = {2023},
}

@inproceedings{Magic3D,
  author       = {Chen{-}Hsuan Lin and
                  Jun Gao and
                  Luming Tang and
                  Towaki Takikawa and
                  Xiaohui Zeng and
                  Xun Huang and
                  Karsten Kreis and
                  Sanja Fidler and
                  Ming{-}Yu Liu and
                  Tsung{-}Yi Lin},
  title        = {Magic3D: High-Resolution Text-to-3D Content Creation},
  booktitle    = {IEEE/CVF Conference on Computer Vision and Pattern Recognition},
  pages        = {300--309},
  year         = {2023},
}

@inproceedings{zero123,
  author       = {Ruoshi Liu and
                  Rundi Wu and
                  Basile Van Hoorick and
                  Pavel Tokmakov and
                  Sergey Zakharov and
                  Carl Vondrick},
  title        = {Zero-1-to-3: Zero-shot One Image to 3D Object},
  booktitle    = {IEEE/CVF International Conference on Computer Vision},
  pages        = {9264--9275},
  year         = {2023},
}

@inproceedings{lgm,
  title={Lgm: Large multi-view gaussian model for high-resolution 3d content creation},
  author={Tang, Jiaxiang and Chen, Zhaoxi and Chen, Xiaokang and Wang, Tengfei and Zeng, Gang and Liu, Ziwei},
  booktitle={European Conference on Computer Vision},
  pages={1--18},
  year={2024},
  organization={Springer}
}

@inproceedings{3dtopia-xl,
  title={3dtopia-xl: Scaling high-quality 3d asset generation via primitive diffusion},
  author={Chen, Zhaoxi and Tang, Jiaxiang and Dong, Yuhao and Cao, Ziang and Hong, Fangzhou and Lan, Yushi and Wang, Tengfei and Xie, Haozhe and Wu, Tong and Saito, Shunsuke and others},
  booktitle={IEEE/CVF Conference on Computer Vision and Pattern Recognition},
  pages={26576--26586},
  year={2025}
}

@inproceedings{trellis,
  title={Structured 3d latents for scalable and versatile 3d generation},
  author={Xiang, Jianfeng and Lv, Zelong and Xu, Sicheng and Deng, Yu and Wang, Ruicheng and Zhang, Bowen and Chen, Dong and Tong, Xin and Yang, Jiaolong},
  booktitle={IEEE/CVF Conference on Computer Vision and Pattern Recognition},
  pages={21469--21480},
  year={2025}
}

@article{hunyuan3d,
    title={Hunyuan3D 2.0: Scaling Diffusion Models for High Resolution Textured 3D Assets Generation},
    author={Tencent Hunyuan3D Team},
    year={2025},
    journal={arXiv preprint arXiv:2501.12202}
}

@inproceedings{reparo,
  title={Reparo: Compositional 3d assets generation with differentiable 3d layout alignment},
  author={Han, Haonan and Yang, Rui and Liao, Huan and Xing, Jiankai and Xu, Zunnan and Yu, Xiaoming and Zha, Junwei and Li, Xiu and Li, Wanhua},
  booktitle={IEEE/CVF International Conference on Computer Vision},
  pages={25367--25377},
  year={2025}
}

@article{sing3d,
  title={Towards geometric and textural consistency 3d scene generation via single image-guided model generation and layout optimization},
  author={Tang, Xiang and Li, Ruotong and Fan, Xiaopeng},
  journal={arXiv preprint arXiv:2507.14841},
  year={2025}
}

@inproceedings{gen3dsr,
  title={Gen3DSR: Generalizable 3d scene reconstruction via divide and conquer from a single view},
  author={Dogaru, Andreea and {\"O}zer, Mert and Egger, Bernhard},
  booktitle={International Conference on 3D Vision 2025},
  year={2025}
}

@article{CAST,

      title={CAST: Component-Aligned 3D Scene Reconstruction from an RGB Image}, 

      author={Kaixin Yao and Longwen Zhang and Xinhao Yan and Yan Zeng and Qixuan Zhang and Lan Xu and Wei Yang and Jiayuan Gu and Jingyi Yu},

      year={2025},

      journal={arXiv preprint arXiv:2502.12894},
      
}

@inproceedings{3dfront,
  title={3d-front: 3d furnished rooms with layouts and semantics},
  author={Fu, Huan and Cai, Bowen and Gao, Lin and Zhang, Ling-Xiao and Wang, Jiaming and Li, Cao and Zeng, Qixun and Sun, Chengyue and Jia, Rongfei and Zhao, Binqiang and others},
  booktitle={IEEE/CVF International Conference on Computer Vision},
  pages={10933--10942},
  year={2021}
}

@article{flux,
  title={FLUX. 1 Kontext: Flow Matching for In-Context Image Generation and Editing in Latent Space},
  author={Labs, Black Forest and Batifol, Stephen and Blattmann, Andreas and Boesel, Frederic and Consul, Saksham and Diagne, Cyril and Dockhorn, Tim and English, Jack and English, Zion and Esser, Patrick and others},
  journal={arXiv preprint arXiv:2506.15742},
  year={2025}
}

@inproceedings{nie2020total3dunderstanding,
  title={Total3dunderstanding: Joint layout, object pose and mesh reconstruction for indoor scenes from a single image},
  author={Nie, Yinyu and Han, Xiaoguang and Guo, Shihui and Zheng, Yujian and Chang, Jian and Zhang, Jian Jun},
  booktitle={IEEE/CVF Conference on Computer Vision and Pattern Recognition},
  pages={55--64},
  year={2020}
}

@article{shuang2025direct3d,
  title={Direct3D: Scalable Image-to-3D Generation via 3D Latent Diffusion Transformer},
  author={Shuang, Wu and Lin, Youtian and Zeng, Yifei and Zhang, Feihu and Xu, Jingxi and Torr, Philip and Cao, Xun and Yao, Yao},
  journal={Advances in Neural Information Processing Systems},
  volume={37},
  pages={121859--121881},
  year={2025}
}

@article{ren2024grounded,
  title={Grounded sam: Assembling open-world models for diverse visual tasks},
  author={Ren, Tianhe and Liu, Shilong and Zeng, Ailing and Lin, Jing and Li, Kunchang and Cao, He and Chen, Jiayu and Huang, Xinyu and Chen, Yukang and Yan, Feng and others},
  journal={arXiv preprint arXiv:2401.14159},
  year={2024}
}

@inproceedings{CLIP,
  author       = {Alec Radford and
                  Jong Wook Kim and
                  Chris Hallacy and
                  Aditya Ramesh and
                  Gabriel Goh and
                  Sandhini Agarwal and
                  Girish Sastry and
                  Amanda Askell and
                  Pamela Mishkin and
                  Jack Clark and
                  Gretchen Krueger and
                  Ilya Sutskever},
  title        = {Learning Transferable Visual Models From Natural Language Supervision},
  booktitle    = {International Conference on Machine Learning},
  volume       = {139},
  pages        = {8748--8763},
  year         = {2021},
}

@inproceedings{dino,
  title={Emerging Properties in Self-Supervised Vision Transformers},
  author={Caron, Mathilde and Touvron, Hugo and Misra, Ishan and J\'egou, Herv\'e  and Mairal, Julien and Bojanowski, Piotr and Joulin, Armand},
  booktitle={IEEE/CVF International Conference on Computer Vision},
  year={2021}
}

@inproceedings{liang2024luciddreamer,
  title={Luciddreamer: Towards high-fidelity text-to-3d generation via interval score matching},
  author={Liang, Yixun and Yang, Xin and Lin, Jiantao and Li, Haodong and Xu, Xiaogang and Chen, Yingcong},
  booktitle={IEEE/CVF Conference on Computer Vision and Pattern Recognition},
  pages={6517--6526},
  year={2024}
}

@inproceedings{ProlificDreamer,
  author       = {Zhengyi Wang and
                  Cheng Lu and
                  Yikai Wang and
                  Fan Bao and
                  Chongxuan Li and
                  Hang Su and
                  Jun Zhu},
  title        = {ProlificDreamer: High-Fidelity and Diverse Text-to-3D Generation with
                  Variational Score Distillation},
  booktitle    = {Advances in Neural Information Processing Systems},
  year         = {2023},
}

@inproceedings{chen2024comboverse,
  title={Comboverse: Compositional 3d assets creation using spatially-aware diffusion guidance},
  author={Chen, Yongwei and Wang, Tengfei and Wu, Tong and Pan, Xingang and Jia, Kui and Liu, Ziwei},
  booktitle={European Conference on Computer Vision},
  pages={128--146},
  year={2024},
  organization={Springer}
}

@article{xu2024instantmesh,
  title={Instantmesh: Efficient 3d mesh generation from a single image with sparse-view large reconstruction models},
  author={Xu, Jiale and Cheng, Weihao and Gao, Yiming and Wang, Xintao and Gao, Shenghua and Shan, Ying},
  journal={arXiv preprint arXiv:2404.07191},
  year={2024}
}

@article{MVDream,
  author = {Shi, Yichun and Wang, Peng and Ye, Jianglong and Mai, Long and Li, Kejie and Yang, Xiao},
  title = {MVDream: Multi-view Diffusion for 3D Generation},
  journal = {arXiv preprint arXiv:2308.16512},
  year = {2023},
}

@inproceedings{zou2024triplane,
  title={Triplane meets gaussian splatting: Fast and generalizable single-view 3d reconstruction with transformers},
  author={Zou, Zi-Xin and Yu, Zhipeng and Guo, Yuan-Chen and Li, Yangguang and Liang, Ding and Cao, Yan-Pei and Zhang, Song-Hai},
  booktitle={IEEE/CVF Conference on Computer Vision and Pattern Recognition},
  pages={10324--10335},
  year={2024}
}

@inproceedings{chu2023buol,
  title={Buol: A bottom-up framework with occupancy-aware lifting for panoptic 3d scene reconstruction from a single image},
  author={Chu, Tao and Zhang, Pan and Liu, Qiong and Wang, Jiaqi},
  booktitle={IEEE/CVF Conference on Computer Vision and Pattern Recognition},
  pages={4937--4946},
  year={2023}
}

@inproceedings{liu2022towards,
  title={Towards high-fidelity single-view holistic reconstruction of indoor scenes},
  author={Liu, Haolin and Zheng, Yujian and Chen, Guanying and Cui, Shuguang and Han, Xiaoguang},
  booktitle={European Conference on Computer Vision},
  pages={429--446},
  year={2022},
  organization={Springer}
}

@article{paschalidou2021atiss,
  title={Atiss: Autoregressive transformers for indoor scene synthesis},
  author={Paschalidou, Despoina and Kar, Amlan and Shugrina, Maria and Kreis, Karsten and Geiger, Andreas and Fidler, Sanja},
  journal={Advances in Neural Information Processing Systems},
  volume={34},
  pages={12013--12026},
  year={2021}
}

@article{gao2024diffcad,
  title={Diffcad: Weakly-supervised probabilistic cad model retrieval and alignment from an rgb image},
  author={Gao, Daoyi and Rozenberszki, D{\'a}vid and Leutenegger, Stefan and Dai, Angela},
  journal={ACM Transactions on Graphics},
  volume={43},
  number={4},
  pages={1--15},
  year={2024},
  publisher={ACM New York, NY, USA}
}

@inproceedings{gumeli2022roca,
  title={Roca: Robust cad model retrieval and alignment from a single image},
  author={G{\"u}meli, Can and Dai, Angela and Nie{\ss}ner, Matthias},
  booktitle={IEEE/CVF Conference on Computer Vision and Pattern Recognition},
  pages={4022--4031},
  year={2022}
}

@inproceedings{kuo2021patch2cad,
  title={Patch2cad: Patchwise embedding learning for in-the-wild shape retrieval from a single image},
  author={Kuo, Weicheng and Angelova, Anelia and Lin, Tsung-Yi and Dai, Angela},
  booktitle={IEEE/CVF International Conference on Computer Vision},
  pages={12589--12599},
  year={2021}
}

@inproceedings{chen2023fantasia3d,
  title={Fantasia3d: Disentangling geometry and appearance for high-quality text-to-3d content creation},
  author={Chen, Rui and Chen, Yongwei and Jiao, Ningxin and Jia, Kui},
  booktitle={IEEE/CVF International Conference on Computer Vision},
  pages={22246--22256},
  year={2023}
}

@inproceedings{DreamView,
  author = {Yan, Junkai and Gao, Yipeng and Yang, Qize and Wei, Xihan and Xie, Xuansong and Wu, Ancong and Zheng, Wei-Shi},
  title = {DreamView: Injecting View-specific Text Guidance into Text-to-3D Generation},
  booktitle = {European Conference on Computer Vision},
  year = {2024}
}

@inproceedings{tang2024diffuscene,
  title={Diffuscene: Denoising diffusion models for generative indoor scene synthesis},
  author={Tang, Jiapeng and Nie, Yinyu and Markhasin, Lev and Dai, Angela and Thies, Justus and Nie{\ss}ner, Matthias},
  booktitle={IEEE/CVF Conference on Computer Vision and Pattern Recognition},
  pages={20507--20518},
  year={2024}
}

@article{rahamim2024lay,
  title={Lay-a-scene: Personalized 3d object arrangement using text-to-image priors},
  author={Rahamim, Ohad and Segev, Hilit and Achituve, Idan and Atzmon, Yuval and Kasten, Yoni and Chechik, Gal},
  journal={arXiv preprint arXiv:2406.00687},
  year={2024}
}

@inproceedings{liu2024one,
  title={One-2-3-45++: Fast single image to 3d objects with consistent multi-view generation and 3d diffusion},
  author={Liu, Minghua and Shi, Ruoxi and Chen, Linghao and Zhang, Zhuoyang and Xu, Chao and Wei, Xinyue and Chen, Hansheng and Zeng, Chong and Gu, Jiayuan and Su, Hao},
  booktitle={IEEE/CVF Conference on Computer Vision and Pattern Recognition},
  pages={10072--10083},
  year={2024}
}

@inproceedings{voleti2024sv3d,
  title={Sv3d: Novel multi-view synthesis and 3d generation from a single image using latent video diffusion},
  author={Voleti, Vikram and Yao, Chun-Han and Boss, Mark and Letts, Adam and Pankratz, David and Tochilkin, Dmitry and Laforte, Christian and Rombach, Robin and Jampani, Varun},
  booktitle={European Conference on Computer Vision},
  pages={439--457},
  year={2024}
}
\bibliographystyle{icml2026}

\newpage
\appendix
\section{Experimental Setup}

\paragraph{Metrics.}  
We employ multi-dimensional evaluation metrics to assess multi-instance generation, focusing on both global and local spatial fidelity. 

\textbf{(i) For Objective Evaluation}, following prior works~\cite{nie2020total3dunderstanding, midi}, we adopt Chamfer Distance (CD) and F-Score (FS), which provide a general measure of reconstruction quality and spatial fidelity at the global scene level (CD-S, FS-S) and local object level (CD-O, FS-O). 
To further capture global layout alignment and instance-level separation, we introduce two complementary metrics. 

Layout Consistency Distance (LCD) evaluates the accuracy of object placement by computing the centroid-based Chamfer Distance between predicted and ground-truth instances:
\begin{align}
\text{LCD} &= 
\frac{1}{|\mathcal{C}_{pred}|} \sum_{x \in \mathcal{C}_{pred}} \min_{y \in \mathcal{C}_{gt}} \|x - y\|_2 \\
&\quad + \frac{1}{|\mathcal{C}_{gt}|} \sum_{y \in \mathcal{C}_{gt}} \min_{x \in \mathcal{C}_{pred}} \|y - x\|_2,
\end{align}
where $\mathcal{C}_{pred}$ and $\mathcal{C}_{gt}$ denote the sets of centroids of predicted and ground-truth object instances, respectively, and $\|\cdot\|_2$ represents the Euclidean distance between two points. The first term measures how close each predicted object is to its nearest ground-truth object, while the second term ensures all ground-truth instances are covered by predictions.

Separation Success Rate (SSR) measures how well the predicted number of independent instances matches the ground truth, reflecting instance distinctiveness regardless of exact shape or position:
\begin{equation}
\text{SSR} = \frac{\min(N_{pred}, N_{gt})}{\max(N_{pred}, N_{gt})},
\end{equation}
where $N_{pred}$ and $N_{gt}$ denote the number of predicted and ground-truth independent instances, respectively.

\textbf{(ii) For Subjective Evaluation}, we conduct a user study where participants perform blind comparative ratings on randomly paired results based on three criteria: Global Layout Alignment, Local Instance Distinctiveness, and Local Instance Fidelity. 

\section{More Experimental Results}

\subsection{Large-Scale Quantitative Evaluation}

Table~\ref{tab:large_scale_eval} reports the quantitative results on a larger test set of 500 samples.
\textbf{(i)} Compared with the base Hunyuan3D 2.0 model, TIMI consistently improves both global spatial fidelity and local instance quality. In particular, TIMI achieves lower LCD and CD-S, indicating more accurate scene-level layout and geometry, while also improving FS-S.
\textbf{(ii)} Compared with MIDI, TIMI obtains better performance across all reported metrics under this larger-scale setting. This suggests that the improvement is not limited to a small set of examples, but remains stable across a more diverse evaluation set.
\textbf{(iii)} The overall trends are consistent under the 95\% confidence intervals, providing stronger evidence that TIMI improves multi-instance spatial fidelity in a reliable manner.

\begin{table}[t]
\centering
\caption{Large-scale quantitative evaluation results. $\pm$ denotes the 95\% confidence interval.}
\label{tab:large_scale_eval}
\setlength{\tabcolsep}{4pt}
\resizebox{\linewidth}{!}{
\begin{tabular}{lccc}
\toprule
Method & LCD $\downarrow$ & CD-S $\downarrow$ & FS-S $\uparrow$ \\
\midrule
DPA           & 0.634$\pm$0.020 & 0.0711$\pm$0.0342 & 0.193$\pm$0.013 \\
Hunyuan3D 2.0 & 0.609$\pm$0.023 & 0.0451$\pm$0.0029 & 0.398$\pm$0.018 \\
MIDI          & 0.608$\pm$0.023 & 0.0494$\pm$0.0031 & 0.324$\pm$0.017 \\
\textbf{TIMI} & \textbf{0.596$\pm$0.023} & \textbf{0.0426$\pm$0.0029} & \textbf{0.410$\pm$0.018} \\
\midrule
Method & SSR $\uparrow$ & CD-O $\downarrow$ & FS-O $\uparrow$ \\
\midrule
DPA           & 0.652$\pm$0.021 & 0.1215$\pm$0.0060 & 0.120$\pm$0.010 \\
Hunyuan3D 2.0 & 0.695$\pm$0.019 & 0.1062$\pm$0.0071 & 0.268$\pm$0.018 \\
MIDI          & 0.690$\pm$0.018 & 0.1129$\pm$0.0062 & 0.189$\pm$0.014 \\
\textbf{TIMI} & \textbf{0.705$\pm$0.020} & \textbf{0.1024$\pm$0.0069} & \textbf{0.274$\pm$0.018} \\
\bottomrule
\end{tabular}
}
\end{table}

\begin{table}[h]
\centering
\caption{Cross-backbone generalization results on Trellis.}
\label{tab:trellis_generalization}
\resizebox{\linewidth}{!}{
\begin{tabular}{lcccccc}
\toprule
Method & LCD $\downarrow$ & CD-S $\downarrow$ & FS-S $\uparrow$ & SSR $\uparrow$ & CD-O $\downarrow$ & FS-O $\uparrow$ \\
\midrule
Trellis        & 0.692 & 0.0668 & 0.221 & 0.776 & 0.0899 & 0.131 \\
Trellis + TIMI & \textbf{0.635} & \textbf{0.0622} & \textbf{0.273} & \textbf{0.785} & \textbf{0.0813} & \textbf{0.156} \\
\bottomrule
\end{tabular}
}
\end{table}

\subsection{Cross-Backbone Generalization}

Table~\ref{tab:trellis_generalization} evaluates the generalizability of TIMI by applying it to the Trellis backbone without modifying the model architecture or performing additional training.
\textbf{(i)} TIMI consistently improves Trellis on all global metrics, reducing LCD from 0.692 to 0.635 and CD-S from 0.0668 to 0.0622, while increasing FS-S from 0.221 to 0.273. This indicates that TIMI can improve scene-level layout and geometry over the original backbone.
\textbf{(ii)} TIMI also improves all local instance-level metrics, including SSR, CD-O, and FS-O, suggesting better instance separation and local object fidelity.
\textbf{(iii)} Although the absolute performance still depends on the capacity of the underlying backbone, the consistent improvements on Trellis demonstrate that TIMI is not restricted to Hunyuan3D 2.0. Instead, it acts as a general training-free guidance mechanism that can be plugged into different I23D backbones.

\end{document}